\documentclass[10pt,journal,compsoc]{IEEEtran}
\usepackage[utf8]{inputenc}
\usepackage{cite}
\usepackage{amsmath,amssymb,amsfonts}
\usepackage{algorithmic}
\usepackage{graphicx}
\usepackage{textcomp}
\usepackage{balance}
\usepackage{graphicx}
\usepackage{amsmath}
\usepackage{amssymb}
\usepackage{setspace}
\usepackage{enumerate}
\usepackage{microtype}
\usepackage{hyperref}
\usepackage{color}
\usepackage{subcaption}
\usepackage[usenames,dvipsnames]{xcolor}
\usepackage{tcolorbox}
\usepackage{tabularx}
\usepackage{array}
\usepackage{colortbl}
\usepackage{multirow}
\newcolumntype{C}[1]{>{\centering\let\newline\\\arraybackslash\hspace{0pt}}m{#1}}

\newtheorem{theorem}{Theorem}

\begin{document}
\title{Automatic Landmark Detection and Registration of Brain Cortical Surfaces via Quasi-Conformal Geometry and Convolutional Neural Networks}
\author{Yuchen Guo, Qiguang Chen, Gary P. T. Choi, Lok Ming Lui
\thanks{This work was supported in part by the National Science Foundation under Grant No. DMS-2002103 (to Gary P. T. Choi), and HKRGC GRF under Project ID 14305919 (to Lok Ming Lui).}
\thanks{Yuchen Guo is with the Department of Mathematics, The Chinese University of Hong Kong (email: ycguo@math.cuhk.edu.hk).}
\thanks{Qiguang Chen is with the Department of Mathematics, The Chinese University of Hong Kong (email: qgchen@math.cuhk.edu.hk).}
\thanks{Gary P. T. Choi is with the Department of Mathematics, Massachusetts Institute of Technology (email: ptchoi@mit.edu).}
\thanks{Lok Ming Lui is with the Department of Mathematics, The Chinese University of Hong Kong (email: lmlui@math.cuhk.edu.hk).}
}

\IEEEtitleabstractindextext{%
\begin{abstract}
In medical imaging, surface registration is extensively used for performing systematic comparisons between anatomical structures, with a prime example being the highly convoluted brain cortical surfaces. To obtain a meaningful registration, a common approach is to identify prominent features on the surfaces and establish a low-distortion mapping between them with the feature correspondence encoded as landmark constraints. Prior registration works have primarily focused on using manually labeled landmarks and solving highly nonlinear optimization problems, which are time-consuming and hence hinder practical applications. In this work, we propose a novel framework for the automatic landmark detection and registration of brain cortical surfaces using quasi-conformal geometry and convolutional neural networks. We first develop a landmark detection network (LD-Net) that allows for the automatic extraction of landmark curves given two prescribed starting and ending points based on the surface geometry. We then utilize the detected landmarks and quasi-conformal theory for achieving the surface registration. Specifically, we develop a coefficient prediction network (CP-Net) for predicting the Beltrami coefficients associated with the desired landmark-based registration and a mapping network called the disk Beltrami solver network (DBS-Net) for generating quasi-conformal mappings from the predicted Beltrami coefficients, with the bijectivity guaranteed by quasi-conformal theory. Experimental results are presented to demonstrate the effectiveness of our proposed framework. Altogether, our work paves a new way for surface-based morphometry and medical shape analysis.
\end{abstract}

}

\maketitle

\IEEEdisplaynontitleabstractindextext

\section{Introduction}
\IEEEPARstart{S}urface registration, the process of finding a 1-1 correspondence between surfaces, has been extensively studied and widely applied to various fields in science, engineering, and medicine. For instance, in medical imaging, surface registration facilitates the comparison between complicated anatomical shapes for disease analysis. To ensure the accuracy of surface registration, landmarks representing salient features on the surfaces are frequently used to guide the registration. For relatively simple shapes like human faces, prominent features such as the eyes can be easily identified as landmarks. However, for more complicated shapes such as the highly convoluted brain cortical surfaces, landmark extraction usually requires manual delineations by medical experts, which makes the task much more time-consuming. Besides the correspondence between the labeled landmarks, it is also essential to find an accurate mapping between the overall surfaces with low distortion, which is traditionally done by solving some optimization problems~\cite{1996A,Fischl1999Cortical,Fischl2008Cortical,yeo2009spherical, robinson2014msm} and is computationally expensive.

Recently, deep learning has emerged as a powerful tool for various complex tasks. To train a suitable model, a large amount of data are usually required. For image registration, it is easy to get millions of images for training and hence convolutional neural networks (CNNs) have achieved huge success. By contrast, surface registration is more complicated as it involves irregular triangular meshes. In particular, due to the lack of data for training, one usually has to augment datasets manually and consider more complicated networks such as graph convolutional networks (GCNs).

\begin{figure}[t!]
    \centering
    \includegraphics[width = 0.85\linewidth]{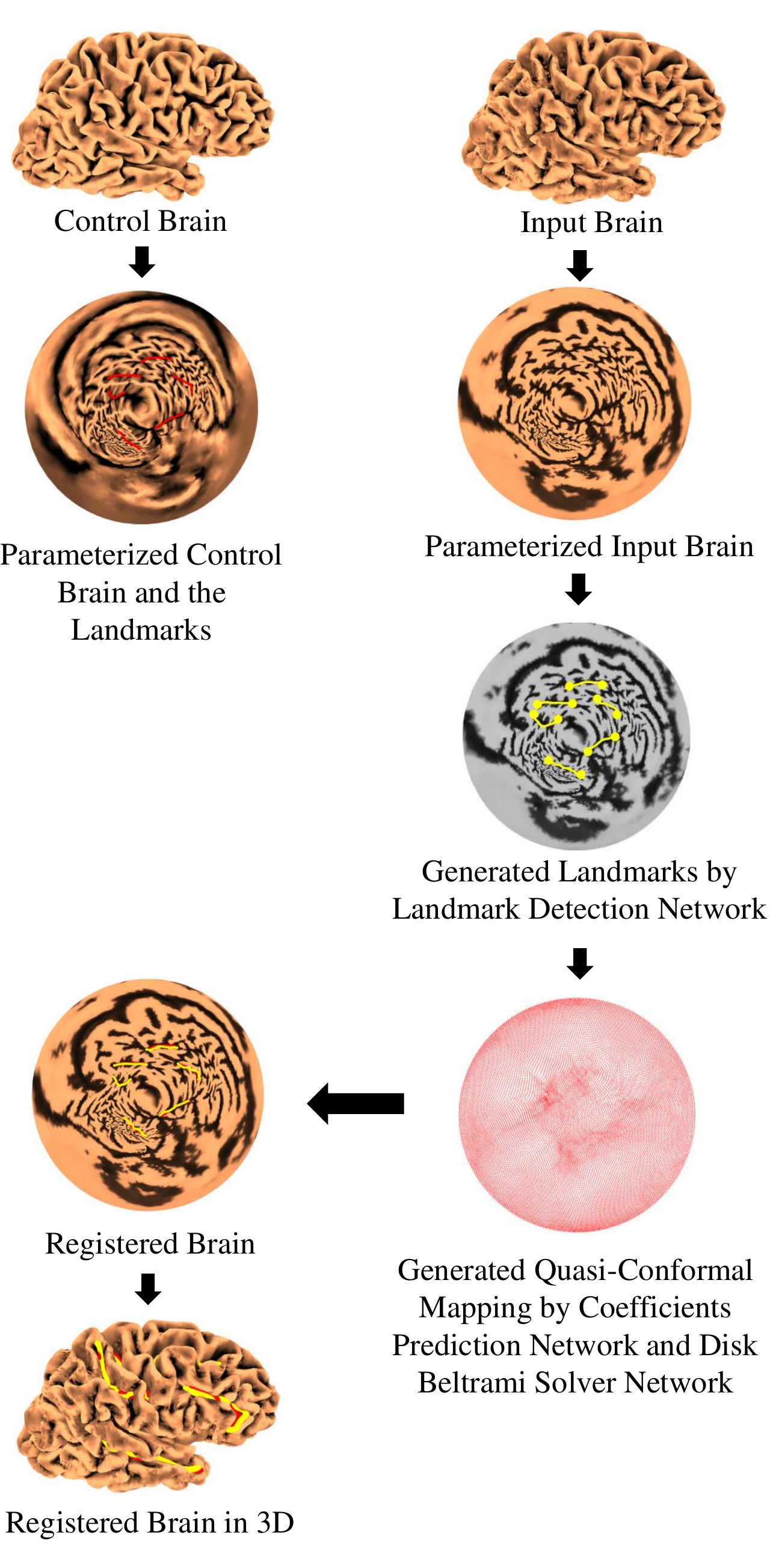}
    \caption{An overview of our proposed framework. Given a control surface with prescribed landmarks, we first conformally parameterize it onto the unit disk. Then, for any given surface, we compute the disk conformal parameterization of it and use the landmark detection network (LD-Net) to extract the feature landmark curves. We then use the coefficient prediction network (CP-Net) and the disk Beltrami solver network (DBS-Net) to register the disk parameterization with that of the control surface, thereby getting the final registration result between the input surface and the control surface.}
    \label{BrainDemo}
\end{figure}

To overcome the above-mentioned issues, in this paper we propose a novel framework for solving the automatic landmark detection and registration problems for brain cortical surfaces by combining quasi-conformal geometry and CNNs (see Fig.~\ref{BrainDemo} for an overview). The contributions of our work are as follows:

\begin{enumerate}[(i)]
    \item We apply quasi-conformal geometry and CNNs for mapping triangular meshes with disk topology.
    \item Feature landmark curves can be automatically extracted from the given surfaces based on only the labeled starting and ending points of them and the surface curvature.
    \item Beltrami coefficients can be automatically generated based on the landmark constraints.
    \item Quasi-conformal mappings can be automatically generated based on the Beltrami coefficients in real time, thereby yielding a landmark-based surface registration.
    \item Once the networks are trained separately, the entire framework can be applied for different input brain surfaces without any additional training.
\end{enumerate}

\section{Related Works}
Surface parameterization and mapping algorithms have been widely studied in recent decades~\cite{floater2005surface,sheffer2007mesh}. In particular, as conformal parameterizations preserve the local geometry well and can be applied to various fields, many conformal parameterization algorithms have been developed by different research groups~\cite{angenent1999conformal,haker2000conformal,desbrun2002intrinsic,levy2002least,2004Cortical,gu2004genus,mullen2008spectral,yueh2017efficient} (see~\cite{lei2021computational} for a survey). 

More recently, quasi-conformal theory has been found useful for surface parameterization with a wide range of applications~\cite{choi2022recent}, such as improving the accuracy of conformal parameterizations and achieving trade-offs between conformality and other prescribed constraints for more flexible parameterizations and registrations. In~\cite{lui2012optimization}, Lui \emph{et~al.} introduced the use of the Beltrami coefficients and established a 1-1 correspondence between Beltrami coefficients and surface diffeomorphisms. The Beltrami coefficients capture essential properties of the mappings and represent them in a simpler space. Based on the Beltrami coefficients, many efficient conformal and quasi-conformal mapping methods have been developed~\cite{lui2014teichmuller,choi2018linear,choi2020parallelizable,choi2021efficient}.

Landmark-based registration, which aims to obtain a 1-1 correspondence between surfaces with prescribed feature landmarks, has been widely studied~\cite{1989Thin,2000Landmark,glaunes2004landmark,2011A,lam2014landmark}. In particular, there have been multiple works on using conformal and quasi-conformal mappings for landmark-based medical shape analysis. For instance, various algorithms for computing landmark-constrained optimized conformal parameterization of brain cortical surfaces have been developed~\cite{wang2005optimization,lui2007landmark,lui2010optimized,choi2015flash}. Landmark-based quasi-conformal registrations have also been used for analyzing teeth~\cite{choi2020tooth,choi2020shape} and human faces~\cite{meng2016tempo,chan2020quasi}.
 
Convolutional neural networks (CNNs), a powerful tool extensively studied and used in recent years, perform excellently in imaging science. In 1989, backpropagation was introduced to the network training process and applied to handwritten digit recognition~\cite{lecun1989backpropagation}. Nowadays, with the improvement of GPUs and advanced network structures such as ResNet~\cite{he2016deep}, researchers are able to train deeper networks to tackle more complicated problems. Besides the improvement of the network structures, large training datasets such as ImageNet~\cite{2009ImageNet} have facilitated the design of more robust networks. More recently, the combination of quasi-conformal geometry and CNNs has been explored~\cite{chen2021learning,law2022quasiconformal}. In particular, Chen \emph{et~al.}~\cite{chen2021learning} developed a network for generating quasi-conformal maps based on any input Beltrami coefficient for rectangular images.

In recent decades, there has been a growing interest in applying networks for surface registration. One common approach is to treat the surface meshes as graphs and use graph neural networks (GNNs)~\cite{ shuman2016vertex,defferrard2016convolutional, niepert2016learning,such2017robust}. Besson \emph{et al.}~\cite{besson2021geometric} used GNNs for analyzing brain meshes and predicting sex and age. Besides, customized kernels have also been used to conduct convolutions on the surfaces. In~\cite{masci2015geodesic, boscaini2016learning}, Masci \emph{et~al.} used kernels defined on Riemannian manifolds for learning shape correspondence. It is also possible to parameterize the surfaces and perform the network learning in the parameter domain~\cite{wu2018registration, zhao2019spherical,seong2018geometric}. Zhao \emph{et al.}~\cite{zhao2021spherical, zhao2021s3reg} parameterized genus-0 cortical surfaces onto the sphere and developed methods for diffeomorphic spherical surface registration using spherical CNNs in both supervised and unsupervised ways. Lin \emph{et~al.}~\cite{lin2020fpconv} proposed to flatten 3D data by a learned kernel and parameterize it onto a regular shape which can be learned by CNNs in 2D. Cheng \emph{et al.}~\cite{cheng2020cortical} developed a learning-based registration framework for cortical surfaces with spherical topology using 2D planar projection and VoxelMorph~\cite{balakrishnan2019voxelmorph}.

The extraction of feature landmarks from complex anatomical shapes is a highly challenging problem. In prior brain cortical surface mapping works, the sulcal landmarks are usually manually delineated~\cite{ono1990atlas,sowell2002mapping,hill2010surface,joshi2012diffeomorphic}. For the automatic extraction of sulci, existing approaches have used pattern recognition system~\cite{mangin2004object,auzias2011diffeomorphic}, supervised learning~\cite{tu2007automated}, geometric algorithm~\cite{kao2007geometric}, geodesic curvature flow~\cite{joshi2012method}, geodesic path density map~\cite{le2012automatic} etc.

\section{Mathematical Background} \label{sec:Background}
\subsection{Quasi-Conformal Theory}
Conformal maps are angle-preserving homeomorphisms between Riemann surfaces. As they may not exist with the presence of landmark constraints in general, we consider a generalization of them with bounded conformality distortion known as quasi-conformal maps. Mathematically, $f : \mathbb{C} \rightarrow \mathbb{C}$ is \emph{quasi-conformal} if it satisfies the Beltrami equation
\begin{equation}
\frac{\partial f}{\partial \bar{z}}=\mu(z) \frac{\partial f}{\partial z},
\label{BeltramiEqu}
\end{equation}
where $\mu:\mathbb{C} \rightarrow \mathbb{C}$ is a complex-valued function called the \emph{Beltrami coefficient} with $\|\mu\|_{\infty}<1$. Infinitesimally, we can also express $f$ by its local parameter around a point $p$:
\begin{equation}
\begin{split}
f(z) &\approx f(p)+f_{z}(p) (z-p) + f_{\bar{z}}(p) \overline{z-p} \\
&= f(p)+f_{z}(p)((z-p)+\mu(p) \overline{z-p}),
\end{split}
\end{equation}
from which we can easily see that $f$ is conformal around a small neighborhood of $p$ if and only if $\mu(p) = 0$. It also shows that $f$ can be expressed as the sum of $f(p)$ and a stretch map $(z-p)+\mu(p) \overline{z-p}$ multiplied by $f_{z}(p)$, and hence it maps infinitesimal circles to infinitesimal ellipses. All the quasi-conformal distortion is caused by $\mu(p)$, which determines the magnitude of stretch or shrinkage of the ellipses. 

Given any Beltrami coefficient $\mu$ with $\|\mu\|_{\infty}<1$, we can find a corresponding quasi-conformal mapping satisfying the Beltrami equation in the distribution sense~\cite{gardiner2000quasiconformal}:
\begin{theorem}
\label{mu2map}
Suppose $\mu: \mathbb{D} \rightarrow \mathbb{C}$ is Lebesgue measurable with $\| \mu \|_{\infty} < 1$. There is a quasi-conformal homeomorphism $\phi$ from $\mathbb{D}$ to itself, which is in the Sobolev space $W^{1,2}(\mathbb{D})$ and satisfies the Beltrami equation~\eqref{BeltramiEqu} in the distribution sense. Furthermore, by fixing 0 and 1, $\phi$ is uniquely determined.
\end{theorem}

Given an orientation preserving homeomorphism $\phi$, we can find the corresponding Beltrami coefficient from~\eqref{BeltramiEqu}:
\begin{equation}
\mu_{\phi}=\frac{\partial \phi}{\partial \bar{z}} / \frac{\partial \phi}{\partial z}.
\end{equation}
The Jacobian $J$ of $\phi$ is related to $\mu_\phi$ as follows:
\begin{equation}
J(\phi)=\left|\frac{\partial \phi}{\partial z}\right|^{2}\left(1-\left|\mu_{\phi}\right|^{2}\right).
\end{equation}
Since $\phi$ is an orientation preserving homeomorphism, $J(\phi)>0$ and $\left|\mu_{\phi}\right|<1$ everywhere. Hence, we must have $\left\|\mu_{\phi}\right\|_{\infty}<1$. 
Theorem \ref{mu2map} indicates that under suitable normalization, every $\mu$ with $\left\|\mu\right\|_{\infty}<1$ is associated with a unique homeomorphism. Therefore, a homeomorphism from $\mathbb{C}$ or $\mathbb{D}$ onto itself can be uniquely determined by its associated Beltrami coefficient.

\subsection{Surface Curvature}
Curvature is an important quantity in differential geometry for assessing how a surface deviates from a plane. 

We define $N : S \rightarrow \mathbb{S}^2 \subseteq \mathbb{R}^3$ to be the \emph{normal map} giving unit vector at each point $p$. Suppose $C$ is a regular curve on $S$, $p$ is a point on $S$ and $k$ is the curvature of $C$ at $p$. We set $\cos \theta = \langle n, N \rangle$, where $N$ is the normal vector to $S$ at $p$ and $n$ is normal to $C$. $k_n = k \cos \theta$ is called the \emph{normal curvature} of $C$ at $p$. The \emph{principal curvatures} at $p$ are the maximum and minimum of the normal curvature, denoted as $k_1$ and $k_2$ respectively. The \emph{mean curvature} at $p$ is defined to be $H = \frac{1}{2}(k_1 + k_2)$.

\subsection{Linear Beltrami Solver (LBS)} \label{part:lbs}
In \cite{lui2013texture}, Lui \emph{et~al.} proposed an efficient method called the \emph{Linear Beltrami Solver} (LBS) to reconstruct the associated quasi-conformal homeomorphism from any given Beltrami coefficient. The method is outlined below.

Let $M_1$ and $M_2$ be two planar domains and $\mu = \rho + i \tau$, where $i^2 = -1$, be a complex-valued function defined on $M_1$. The LBS method aims to reconstruct the homeomorphism $f : M_1 \rightarrow M_2$ associated with the Beltrami coefficient $\mu$. Let $f = u + iv$. From the Beltrami equation \eqref{BeltramiEqu}, we have
\begin{equation}
\mu(f)=\frac{\left(u_{x}-v_{y}\right)+i\left(v_{x}+u_{y}\right)}{\left(u_{x}+v_{y}\right)+i\left(v_{x}-u_{y}\right)},
\end{equation}
where $\alpha_{1}=\frac{(\rho-1)^{2}+\tau^{2}}{1-\rho^{2}-\tau^{2}}, \alpha_{2}=-\frac{2 \tau}{1-\rho^{2}-\tau^{2}}, \alpha_{3}=\frac{(\rho+1)^{2}+\tau^{2}}{1-\rho^{2}-\tau^{2}}$. We have
\begin{equation}
\begin{aligned}
v_{y} &=\alpha_{1} u_{x}+\alpha_{2} u_{y},  \\
-v_{x} &=\alpha_{2} u_{x}+\alpha_{3} u_{y},
\end{aligned}
\quad\text{ and }\quad
\begin{aligned}
-u_{y} &=\alpha_{1} v_{x}+\alpha_{2} v_{y},\\
u_{x} &=\alpha_{2} v_{x}+\alpha_{3} v_{y}.
\end{aligned}
\label{eq:311}
\end{equation}
Since $\nabla \cdot\left(\begin{array}{c}-v_{y} \\ v_{x}\end{array}\right)=0$, we have
\begin{equation}
\nabla \cdot\left(A\left(\begin{array}{c}
u_{x} \\
u_{y}
\end{array}\right)\right)=0 \text { and } \nabla \cdot\left(A\left(\begin{array}{l}
v_{x} \\
v_{y}
\end{array}\right)\right)=0,
\label{LBS}
\end{equation}
where $A=\left(\begin{array}{cc}
\alpha_{1} & \alpha_{2} \\
\alpha_{2} & \alpha_{3}
\end{array}\right)$. By solving the above equations with certain prescribed boundary constraints, the map $f$ can be obtained.

In the discrete case where $M_1$ is a triangular mesh, we need to restrict that $f$ is piecewise linear on each triangular face $T$, which can be written as
\begin{equation} \label{mapping}
f|_{T}(x,y)=
\begin{bmatrix}
u|_{T}(x,y)\\
v|_{T}(x,y)
\end{bmatrix}=\begin{bmatrix}
a_{T}x+b_{T}y+r_{T}\\
c_{T}x+d_{T}y+s_{T}
\end{bmatrix}.
\end{equation}
Hence, the partial derivatives of $f$ at each face $T$ can be denoted as $D_x f(T) = a_T+i c_T$ and $D_y f(T) = b_T+id_T$. Now the gradient $\nabla_T f:=(D_x f(T),D_y f(T))^t$ on $T$ can be computed by solving
\begin{equation}\label{gradient_eqn}
    \left(\begin{array}{c}
            \vec{v_1}-\vec{v_0}  \\
            \vec{v_2}-\vec{v_0} 
    \end{array}\right)
    \nabla_T f = 
    \left(\begin{array}{c}
            f(\vec{v_1})-f(\vec{v_0})  \\
            f(\vec{v_2})-f(\vec{v_0}) 
    \end{array}\right),
\end{equation}
where $[\vec{v_0}, \vec{v_1}]$ and $[\vec{v_0}, \vec{v_2}]$ are two edges on $T$.

The Beltrami coefficient $\mu$ is also discretized on the triangular faces. Denote the discretized functions $\alpha_1, \alpha_2, \alpha_3$ on a face $T$ by $\alpha_1^T, \alpha_2^T, \alpha_3^T$. From \eqref{eq:311}, we have
\begin{equation} \label{eqn3}
\begin{array}{rl}
    -d_{T} &= \alpha_1^T a_{T} + \alpha_2^T b_{T},\\
    c_{T} &= \alpha_2^T a_{T} + \alpha_3^T b_{T},
\end{array}
\text{ and }
\begin{array}{rl}
    -b_{T} &= \alpha_1^T c_{T} + \alpha_2^T d_{T},\\
    a_{T} &= \alpha_2^T c_{T} + \alpha_3^T d_{T}.
\end{array}
\end{equation}

Let $T=[\vec{v_i}, \vec{v_j}, \vec{v_k}]$ and $\vec{w_I}=f(\vec{v_I})$, where $I = i,j,k$. Suppose $v_I = g_I+ih_I$ and $w_I = s_I+it_I$. From \eqref{gradient_eqn}, we have
\begin{equation}
\begin{bmatrix}
a_{T} & b_{T}\\
c_{T} & d_{T}
\end{bmatrix}
\begin{bmatrix}
g_{j}-g_{i} & g_{k}-g_{i}\\
h_{j}-h_{i} & h_{k}-h_{i}
\end{bmatrix}=\begin{bmatrix}
s_{j}-s_{i} & s_{k}-s_{i}\\
t_{j}-t_{i} & t_{k}-t_{i}
\end{bmatrix}.
\end{equation}
Thus,
\begin{equation} \label{derivative}
    \resizebox{.48\textwidth}{!}{$
    \begin{split}
\begin{bmatrix}
a_{T} &   b_{T}\\
 c_{T} &  d_{T}
\end{bmatrix} & = 
 \begin{bmatrix}
 A_{i}^{T}s_{i}+A_{j}^{T}s_{j}+A_{k}^{T}s_{k} &  B_{i}^{T}s_{i}+B_{j}^{T}s_{j}+B_{k}^{T}s_{k}\\
 A_{i}^{T}t_{i}+A_{j}^{T}t_{j}+A_{k}^{T}t_{k} &  B_{i}^{T}t_{i}+B_{j}^{T}t_{j}+B_{k}^{T}t_{k} 
 \end{bmatrix},
    \end{split}$
}
 \end{equation}
where
\begin{equation}
    \resizebox{.48\textwidth}{!}{$
\begin{array}{c}
    A_{i}^{T}=\left(h_{j}-h_{k}\right)/2\cdot \text{Area}(T),\quad B_{i}^{T}=\left(g_{k}-g_{j}\right)/2\cdot \text{Area}(T),\\
    A_{j}^{T}=\left(h_{k}-h_{i}\right)/2\cdot \text{Area}(T),\quad B_{j}^{T}=\left(g_{i}-g_{k}\right)/2\cdot \text{Area}(T),\\
    A_{k}^{T}=\left(h_{i}-h_{j}\right)/2\cdot \text{Area}(T),\quad B_{k}^{T}=\left(g_{j}-g_{i}\right)/2\cdot \text{Area}(T).
\end{array}$
}
\end{equation}
For each vertex $v_{i}$, let $\mathcal{N}_{i}$ be the collection of neighboring faces of $v_{i}$. We can see that
\begin{equation} \label{eqn5}
\sum_{T\in \mathcal{N}_{i}} A_{i}^{T}b_{T} = \sum_{T\in \mathcal{N}_{i}} B_{i}^{T}a_{T}; \quad
\sum_{T\in \mathcal{N}_{i}} A_{i}^{T}d_{T} = \sum_{T\in \mathcal{N}_{i}} B_{i}^{T}c_{T}.
\end{equation}
Substituting \eqref{eqn3} into \eqref{eqn5}, we obtain the following equations:
\begin{equation} \label{linear_system_x}
\sum_{T\in \mathcal{N}_{i}} ( A_{i}^{T} [\alpha_1^T a_{T} + \alpha_2^T b_{T}] + B_{i}^{T}[\alpha_2^T a_{T} + \alpha_3^T b_{T}] ) = 0,
\end{equation}
\begin{equation} \label{linear_system_y}
\sum_{T\in \mathcal{N}_{i}} ( A_{i}^{T} [\alpha_1^T c_{T} + \alpha_2^T d_{T}] + B_{i}^{T}[\alpha_2^T c_{T} + \alpha_3^T d_{T}] ) = 0.
\end{equation}
Therefore, we can solve this linear system to get the $xy$-coordinates and hence the desired mapping $f$.

\subsection{Compression by Fourier Approximation}
Consider the real and imaginary parts of the Beltrami coefficient $\mu$ as two channels of the image and denote them by $\mu_R$ and $\mu_I$ respectively. The Fourier transform of $\mu_j$ (where $j = R$ or $I$) can be expressed as
\begin{equation}
\label{FFT}
\hat{\mu_{j}}(m, n)=\frac{1}{N^{2}} \sum_{k=0}^{N-1} \sum_{l=0}^{N-1} \mu_{j}(k, l) e^{-i \frac{2 \pi k m}{N}} e^{-i\frac{2 \pi l n}{N}}.
\end{equation}
The inverse Fourier transform of $\hat{\mu}_j$ can be written as
\begin{equation}
\mu_j(p, q) = \sum_{m = 0}^{N - 1} \sum_{n = 0}^{N - 1} \hat{\mu}_j(m, n) e ^{i \frac{2\pi p m}{N}} e^{i\frac{2 \pi q n}{N}}.
\end{equation}

In \cite{lui2013texture}, Lui \emph{et~al.} showed that it is possible to compress the Beltrami coefficients by Fourier approximation while preserving the information of homeomorphism and bijectivity. Therefore, keeping only the low-frequency components is sufficient for capturing the majority of deformation in our problem. 

\section{Proposed framework}
In this section, we describe our proposed framework for the automatic landmark detection and surface registration in detail. 

\begin{figure}[t]
    \centering
    \includegraphics[width = \linewidth]{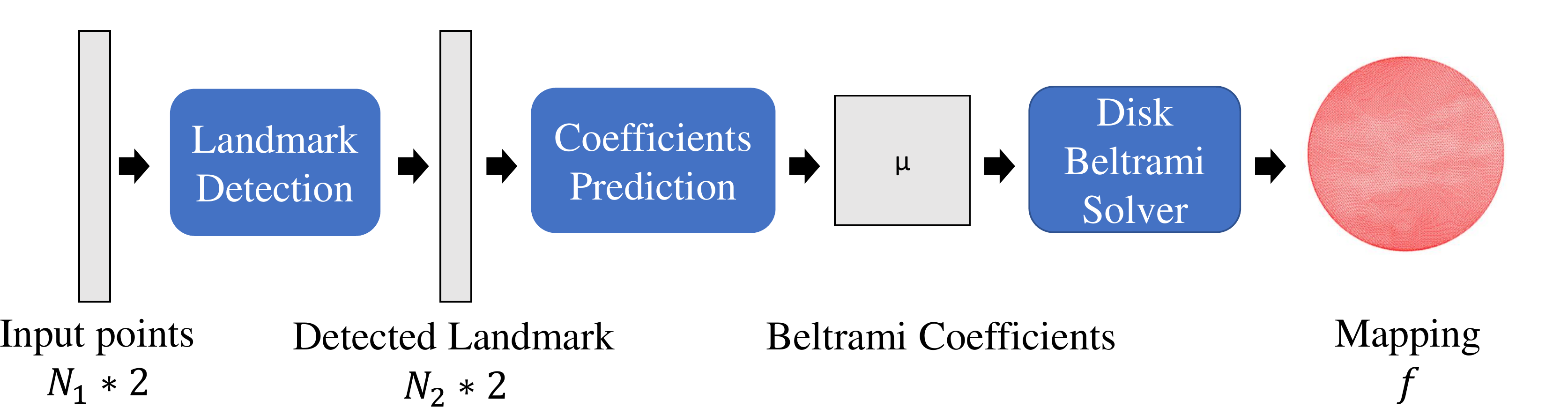}
    \caption{The architecture of the three proposed networks for the automatic landmark detection and surface registration. The network takes the coordinates of $N$ pairs of endpoints of the sulci as the input and generates a landmark-constrained quasi-conformal mapping on the unit disk as the output.}
    \label{Whole}
\end{figure}

Suppose the given brain cortical surfaces are simply-connected open surfaces, i.e. with disk topology. Our strategy is to map them onto the unit disk using conformal parameterization~\cite{choi2015fast}, and then design three networks for achieving the automatic landmark detection and registration (see Fig.~\ref{Whole} for an outline): the landmark detection network (LD-Net), the coefficient prediction network (CP-Net), and the disk Beltrami solver network (DBS-Net). For the landmark detection, we first label the starting and ending points of the desired feature curves as the input of the LD-Net, which extracts the complete landmark curves based on the mean curvature of the input surface. Then, we take the output from the LD-Net as the input of the CP-Net for generating a Beltrami coefficient $\mu$. Finally, we obtain a quasi-conformal mapping associated with the Beltrami coefficient $\mu$ using the DBS-Net. The architecture of each network is introduced in the following subsections.

\subsection{Landmark Detection}
In this section, we describe our proposed landmark detection network (LD-Net). The aim of LD-Net is to take the starting and ending points of the desired feature curves as the input and automatically generate the complete landmark curves along prominent features based on the mean curvature of each surface. 

For each input surface, we compute its mean curvature at every vertex. We then apply the disk conformal mapping method~\cite{choi2015fast} to parameterize the surface onto the unit disk. By representing the mean curvature values on the unit disk and rescaling them to $[0, 255]$, the input surface can be visualized as a single channel 2D image, where the dark parts correspond to the sulci of a cortical surface and the bright parts correspond to the gyri.

\begin{figure}[t]
    \centering
    \includegraphics[width = 0.95\linewidth]{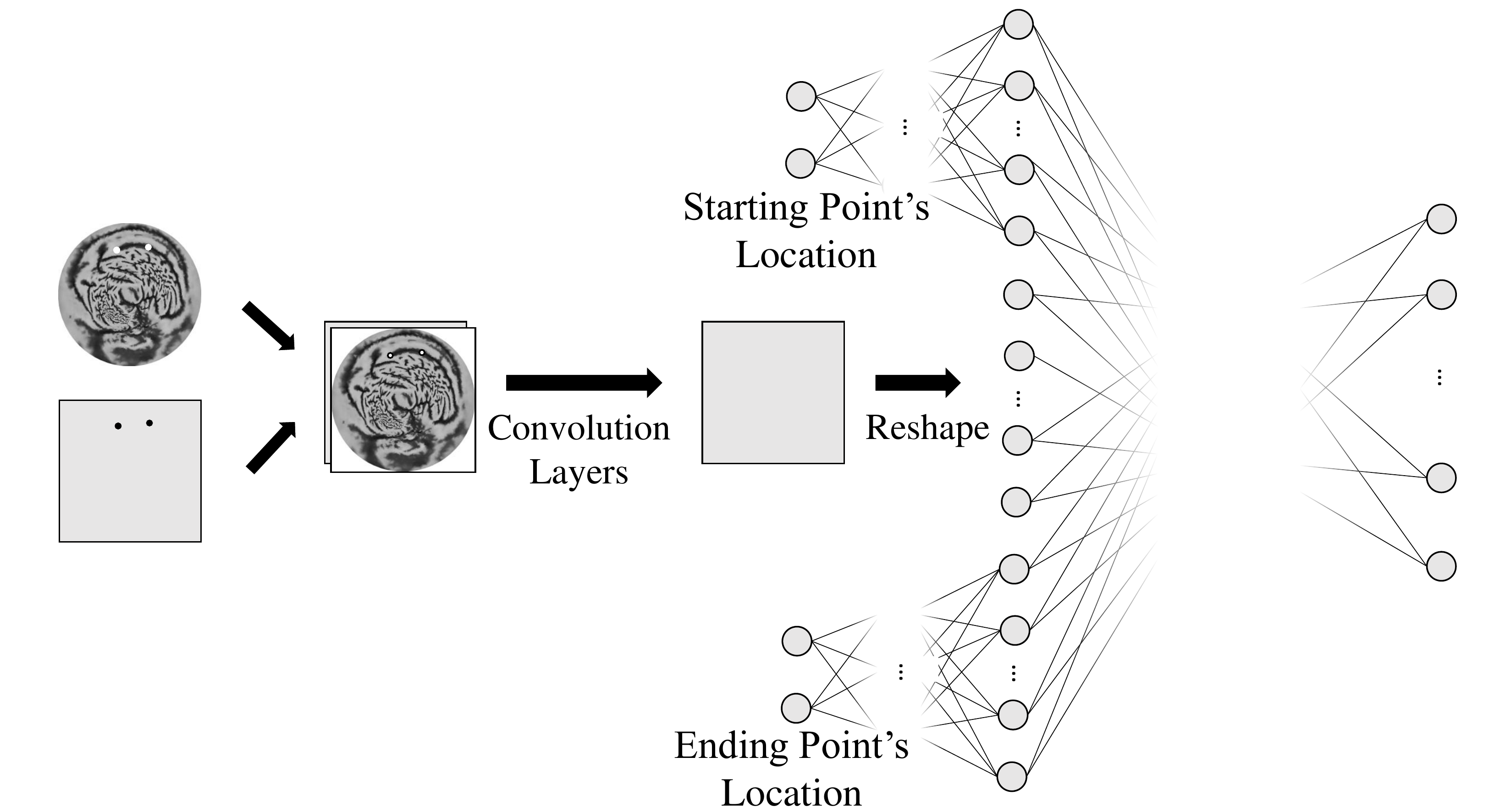}
    \caption{The structure of the LD-Net. We first produce a single channel image of the input surface using conformal parameterization, with the pixel values determined using the mean curvature of the surface. We then label the endpoints of the desired feature curves on the disk and record them in a new channel. The network takes the concatenated image formed by the two above channels as an input, which passes through the convolution layers. It further takes the locations of the starting and ending points as the second and the third input of the network, which separately pass through two MLPs.}
    \label{LandmarkDetect}
\end{figure}

After getting the disk image, we label the location of the endpoints on it and record their corresponding $xy$-coordinates. To ensure that the network can get the location of the points, we create a new channel with the pixel values at the recorded points to be 255 and set the values at all other pixels to be 0. We concatenate this new channel with the disk image together as the first input of the network, which passes through the convolution layers (see Fig.~\ref{LandmarkDetect}). We further take the locations of the starting and ending points as the second and the third input of the network, which separately passes through two multilayer perceptrons (MLPs). To explain this, note that the output inner landmark points should be generated based on all information from the image. Also, taking the coordinates of the starting and ending points as the second and third input effectively forces the inner points to be processed together with the starting and ending locations, so that the output locations are not far from the starting and ending points. Thus, we concatenate the output from these routes and finally get a set of landmark points representing the predicted landmark curves.

\textbf{Loss Function}: For the training of the network, we adopt supervised learning by generating and augmenting data. The loss function for this network is as follows:
\begin{equation}
    \mathcal{L}_{\text{LD}} = \sum_{i=0}^N\|x_i - \hat{x}_i\|^2,
\end{equation}
where $N$ is the number of output points for the prediction of the landmark curves, and $x_i, \hat{x}_i$ are respectively the actual point and the output point from the network for $i = 1, \dots, N$. 

\subsection{Landmark-based Quasi-Conformal Mapping}
As described previously, quasi-conformal mappings can be represented by Beltrami coefficients, and it is easier to generate a 1-1 mapping from the Beltrami coefficients. Therefore, the two main tasks here are: 1) to generate a quasi-conformal mapping based on any input Beltrami coefficient in a fast and robust way, and 2) to generate a Beltrami coefficient that corresponds to a landmark-based registration to be fed into the mapping method.

\subsubsection{Disk Beltrami Solver Network (DBS-Net)}
Note that the LBS method~\cite{lui2013texture} involves solving linear systems and so the computation may be time-consuming if one has to handle a large set of dense triangular meshes. To overcome this problem, Chen \emph{et~al.}~\cite{chen2021learning} proposed a method to train a network that can generate quasi-conformal mappings from Beltrami coefficients more efficiently. However, their method is only applicable to image registration with the data represented using regular grids. To apply this idea for the parameterized irregular triangular meshes in our case, it is necessary to develop a new method.

\begin{figure}[t]
    \centering
    \includegraphics[width = 0.9\linewidth]{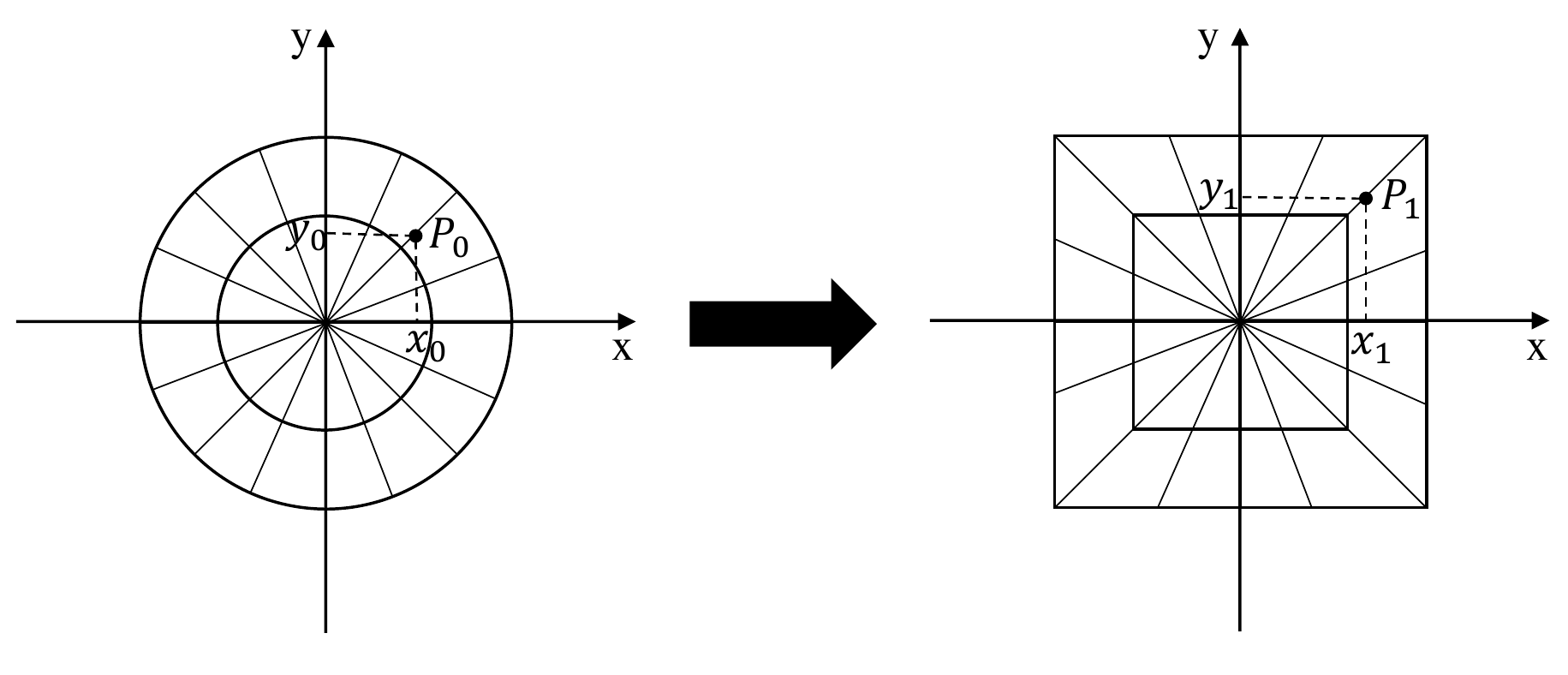}
    \caption{The circle-to-square transformation in our framework.}
    \label{Cir2Rec}
\end{figure}

Here we propose the \emph{Disk Beltrami Solver Network} (DBS-Net) for generating a quasi-conformal mapping based on any input Beltrami coefficient on the disk. To make use of CNNs in the disk, we need to reshape the disk Beltrami coefficients to a square domain. Here, we consider the following transformation:
\begin{equation}
x_1 = \frac{x_0}{\max\{\cos\theta, \sin\theta\}}, \quad y_1 = \frac{y_0}{\max\{\cos\theta, \sin\theta\}},
\end{equation}
where $(x_0, y_0)$, $(x_1, y_1)$ are the coordinates of a point in the disk and its corresponding point in the square, and $\theta$ is the angle of the point in the polar coordinate system (see Fig.~\ref{Cir2Rec}).
 
\begin{figure}[t]
    \centering
    \includegraphics[width = 0.95\linewidth]{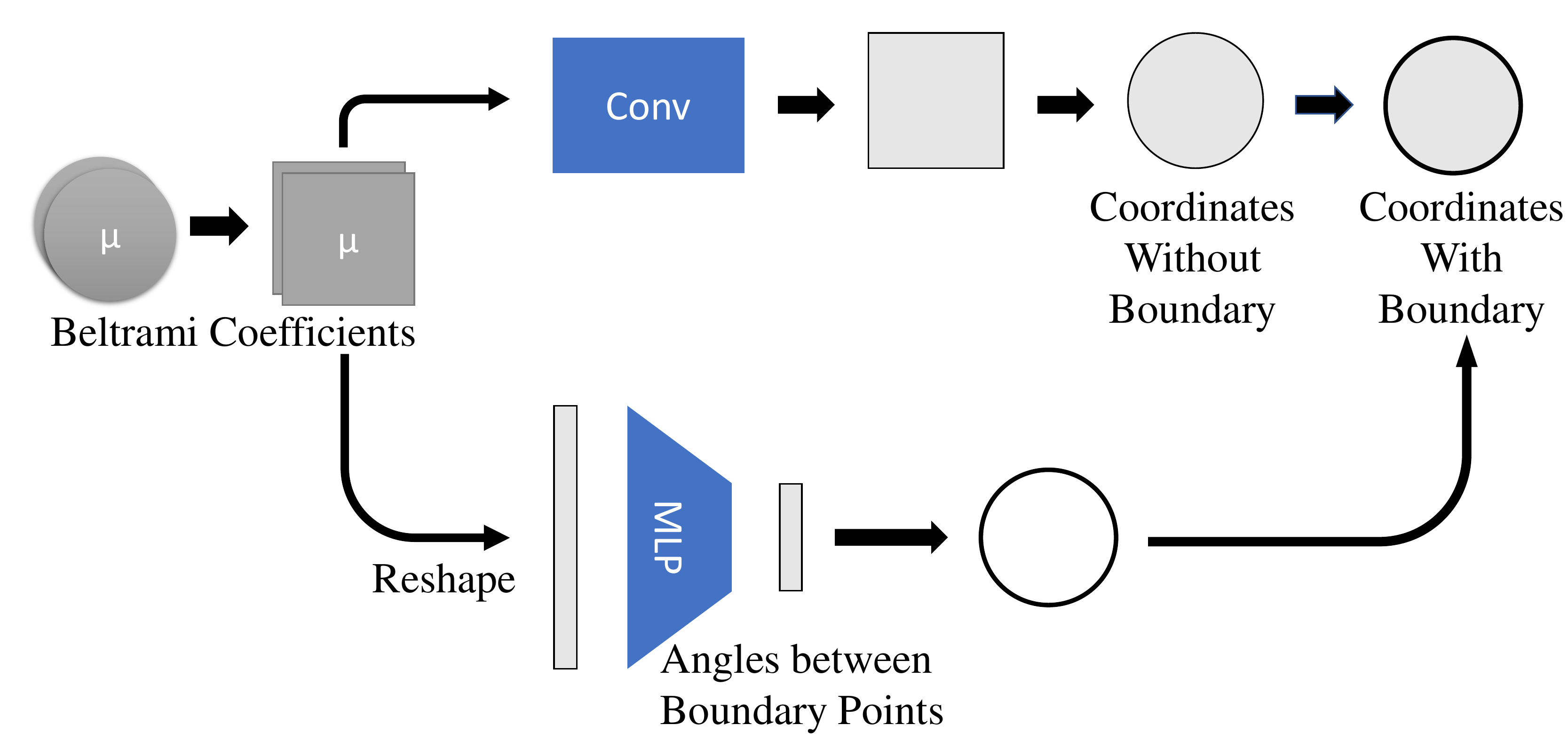}
    \caption{The flow of the proposed DBS-Net.}
    \label{DiskBS}
\end{figure}

After transforming the disk Beltrami coefficients, the DBS-Net uses two routes to determine the inner and boundary coordinates for the quasi-conformal map separately (see  Fig.~\ref{DiskBS}).

\begin{figure}[t]
    \centering
    \includegraphics[width = 0.95\linewidth]{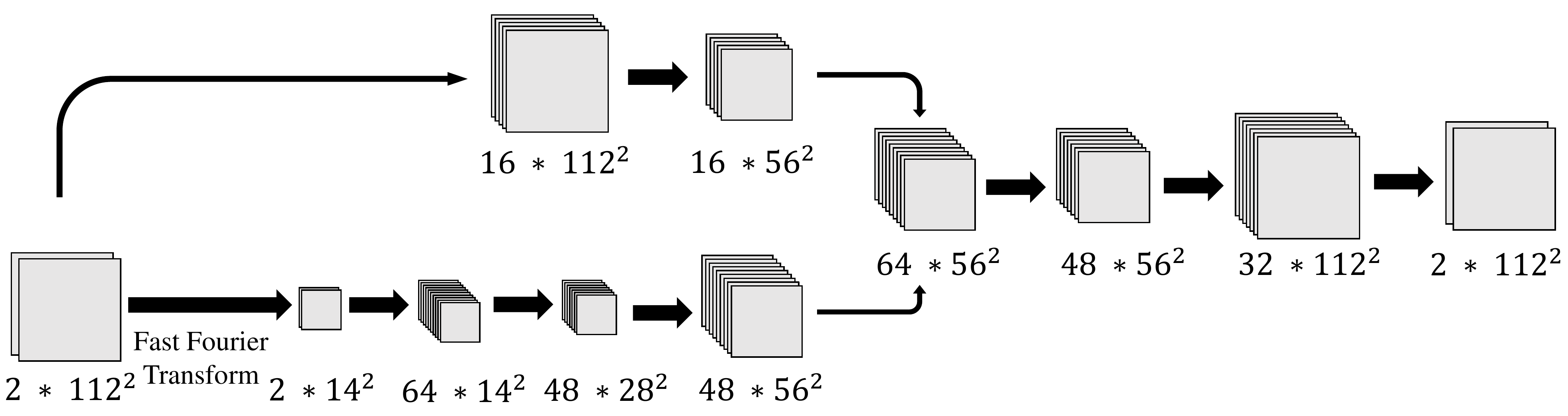}
    \caption{The convolution for generating the inner coordinates of the quasi-conformal maps.}
    \label{InnerCoordinates}
\end{figure}

\textbf{Route for the Inner Coordinates}: Recall that we can use the Fourier approximation to get the majority of information from the Beltrami coefficient $\mu$ by keeping its low-frequency components. Therefore, we first add a fast Fourier transform (FFT) layer to accelerate the training process. To apply \eqref{FFT}, we use two matrices to imitate the computation of the Fourier approximation:
\begin{equation}
\hat{\mu} = M \mu N = (\mu^T M ^ T)^T N,
\end{equation}
where $M$ and $N$ are learnable. Using this process, we can extract the features from the FFT layer and perform convolution to get the output from them. Since the low-frequency parts contain the majority of information, we only need to use bilinear interpolation to upsample and conduct convolution right after the interpolation (see Fig.~\ref{InnerCoordinates}).

While the Fourier approximation can preserve the majority of information, some details will still be lost. To overcome this problem, here we use a shortcut to preserve the detailed information and concatenate the feature map generated from the shortcut with the one that was processed after the Fourier approximation. This allows us to process the data in a more efficient way while also having the details in the output.

\textbf{Route for the Boundary Coordinates}: To reconstruct a quasi-conformal map from a Beltrami coefficient, proper boundary conditions are needed. In our case, we need to restrict the boundary points of the quasi-conformal map to remain on the boundary of the unit disk. It is noteworthy that the boundary points are not meant to be fixed. Instead, we only fix one boundary point to remove the freedom of the global rotation and allow all the remaining boundary points to move along the disk boundary to get a low-distortion mapping. To achieve this, here we design a separate route for determining the boundary coordinates. 

\begin{figure}[t]
    \centering
    \includegraphics[width = \linewidth]{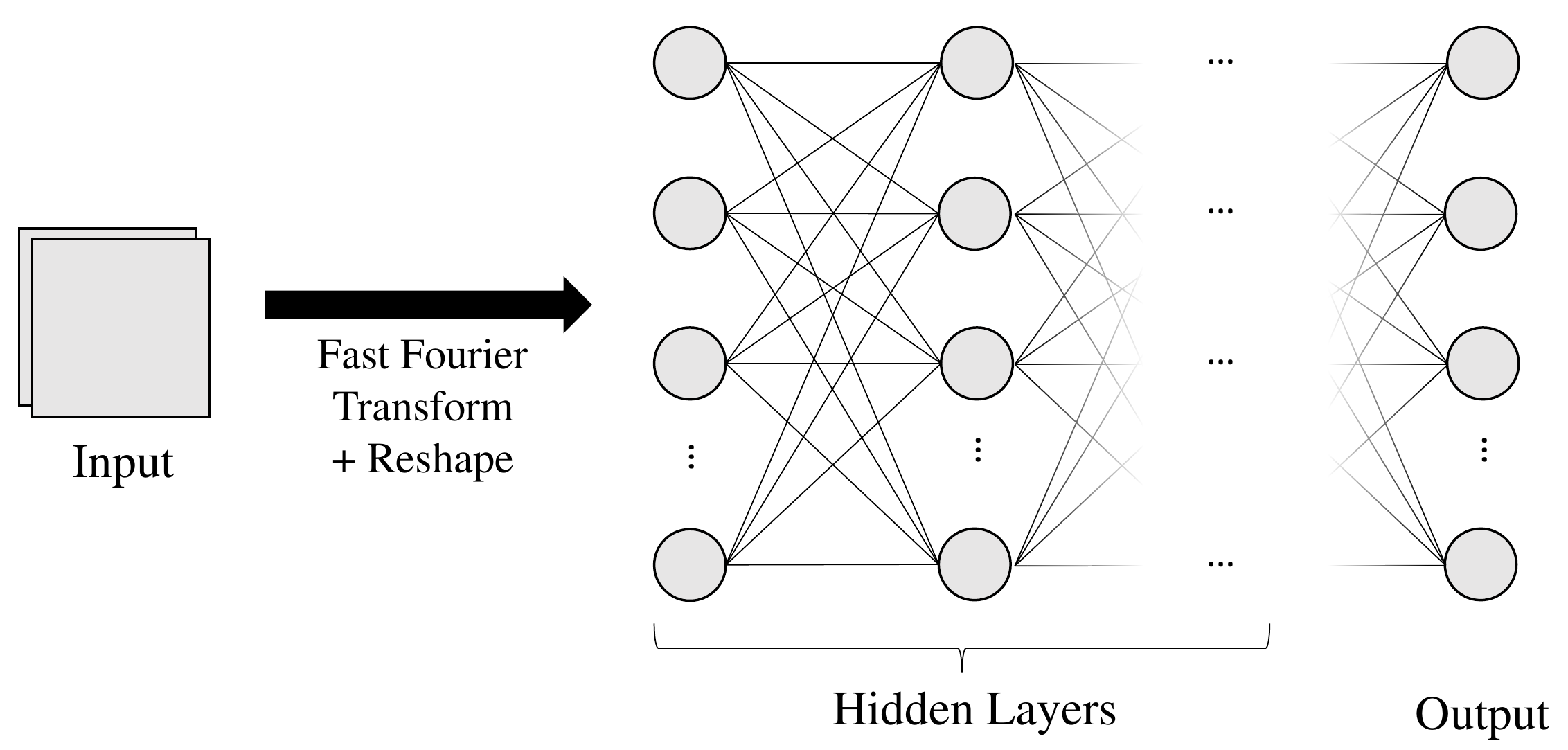}
    \caption{The process of obtaining the boundary coordinates of the quasi-conformal maps.}
    \label{BoundTheta}
\end{figure}

More specifically, we consider the angles between different boundary points. Since the calculation of each point involves its neighboring points and hence is related to every point on the disk, we need the information of the Beltrami coefficient $\mu$ on every face. Therefore, we reshape the input $\mu$ to be a column vector after using the FFT and only keep the low-frequency data. Then, we adopt the MLP to generate the desired angles, thereby determining the boundary coordinates (see Fig.~\ref{BoundTheta}).

\textbf{Loss Function}: For the training, we use an unsupervised setting by adopting LBS in computing the loss function. 

We first consider replacing $a_T$, $b_T$, $c_T$, $d_T$ in~\eqref{linear_system_x} and~\eqref{linear_system_y} with the expression in~\eqref{derivative} to derive the coefficients for every vertex $v_i$ and all vertices $v_l$ adjacent to $v_i$. We have
\begin{equation}
\label{central_coeff}
c_{i}=\sum_{T \in \mathcal{N}_{i}}\left[\alpha_{1}^{T}\left(A_{i}^{T}\right)^{2}+2 \alpha_{2}^{T} A_{i}^{T} B_{i}^{T}+\alpha_{3}^{T}\left(B_{i}^{T}\right)^{2}\right],
\end{equation}
where $\mathcal{N}_{i}$ is the collection of neighboring faces of $v_i$. Note that each edge is shared by two faces. Denoting the two faces sharing the edge $[v_i,v_l]$ by $T_1, T_2$ and the remaining vertices in $T_1$ and $T_2$ by $v_j$ and $v_k$ respectively, the coefficient $c_l$ can be expressed as:
\begin{equation}
\label{cv_coeff}
\begin{aligned}
c_{l}=& \alpha_{1}^{T_{1}} A_{i}^{T_{1}} A_{j}^{T_{1}}+\alpha_{2}^{T_{1}}\left(A_{i}^{T_{1}} B_{j}^{T_{1}}+A_{j}^{T_{1}} B_{i}^{T_{1}}\right)+\alpha_{3}^{T_{1}} B_{i}^{T_{1}} B_{j}^{T_{1}}+\\
& \alpha_{1}^{T_{2}} A_{i}^{T_{2}} A_{k}^{T_{2}}+\alpha_{2}^{T_{2}}\left(A_{i}^{T_{2}} B_{k}^{T_{2}}+A_{k}^{T_{2}} B_{i}^{T_{2}}\right)+\alpha_{3}^{T_{2}} B_{i}^{T_{2}} B_{k}^{T_{2}}.
\end{aligned}
\end{equation}
Therefore, we define the following loss function to evaluate the difference between the generated mapping result and the result produced by LBS:
\begin{equation}
\label{LBS_loss}
\mathcal{L}_{\text{LBS}} = \frac{1}{2N^2}\sum_i (|c_i s_i + \sum_{v_l \in \mathcal{V}_i}c_l s_l| + |c_i t_i + \sum_{v_l \in \mathcal{V}_i}c_l t_l|),
\end{equation}
where $s_i$ and $t_i$ are $x$ and $y$ coordinates of $v_i$ respectively, $N$ is the total number of vertices, and $\mathcal{V}_{i}$ is the collection of all vertices adjacent to $v_i$. This effectively ensures that the output coordinates satisfy~\eqref{linear_system_x} and \eqref{linear_system_y}. 

Moreover, note that the quasi-conformality of a mapping does not change under compositions with conformal maps. With such flexibility, the vertices in the output mapping tend to cluster around the fixed boundary point. To overcome this problem and let the boundary points distribute more evenly on the disk boundary, we introduce another loss function:
\begin{equation}
\mathcal{L}_{\text{Boundary}} = \sum^{N}_{i=1}\frac{1}{\theta_i},
\end{equation}
where $N$ is the number of boundary points and $\theta_i$ is the angle between two neighboring boundary points with $\sum^{N}_{i=1}\theta_i = 2\pi$. Intuitively, each $\frac{1}{\theta_i}$ term effectively prevents any two neighboring boundary points from being too close to each other, thereby ensuring that the boundary points will be distributed more evenly over the entire disk boundary.

Altogether, we propose the following combined loss function:
\begin{equation}
\label{BSnetLoss}
    \mathcal{L}_{\text{DBSNet}} = \eta \mathcal{L}_{\text{LBS}} + \rho \mathcal{L}_{\text{Boundary}},
\end{equation}
where $\eta, \rho$ are two nonnegative balancing parameters. Note that if $\rho$ is too small, $\mathcal{L}_{\text{DBSNet}}$ will be dominated by the first term and the clustering effect will still be present in the mapping result. By contrast, if $\rho$ is too large, the second term in $\mathcal{L}_{\text{DBSNet}}$ will significantly constrain the boundary points. In practice, we find that $\eta = 1$ and $\rho = 10^{-8}$ give satisfactory results. 

\subsubsection{Coefficient Prediction Network (CP-Net)}
The goal of CP-Net is to take the starting points as the input and output a Beltrami coefficient $\mu$, which can then be used for generating the quasi-conformal mapping using the DBS-Net.

\begin{figure}[t]
    \centering
    \includegraphics[width = 0.95\linewidth]{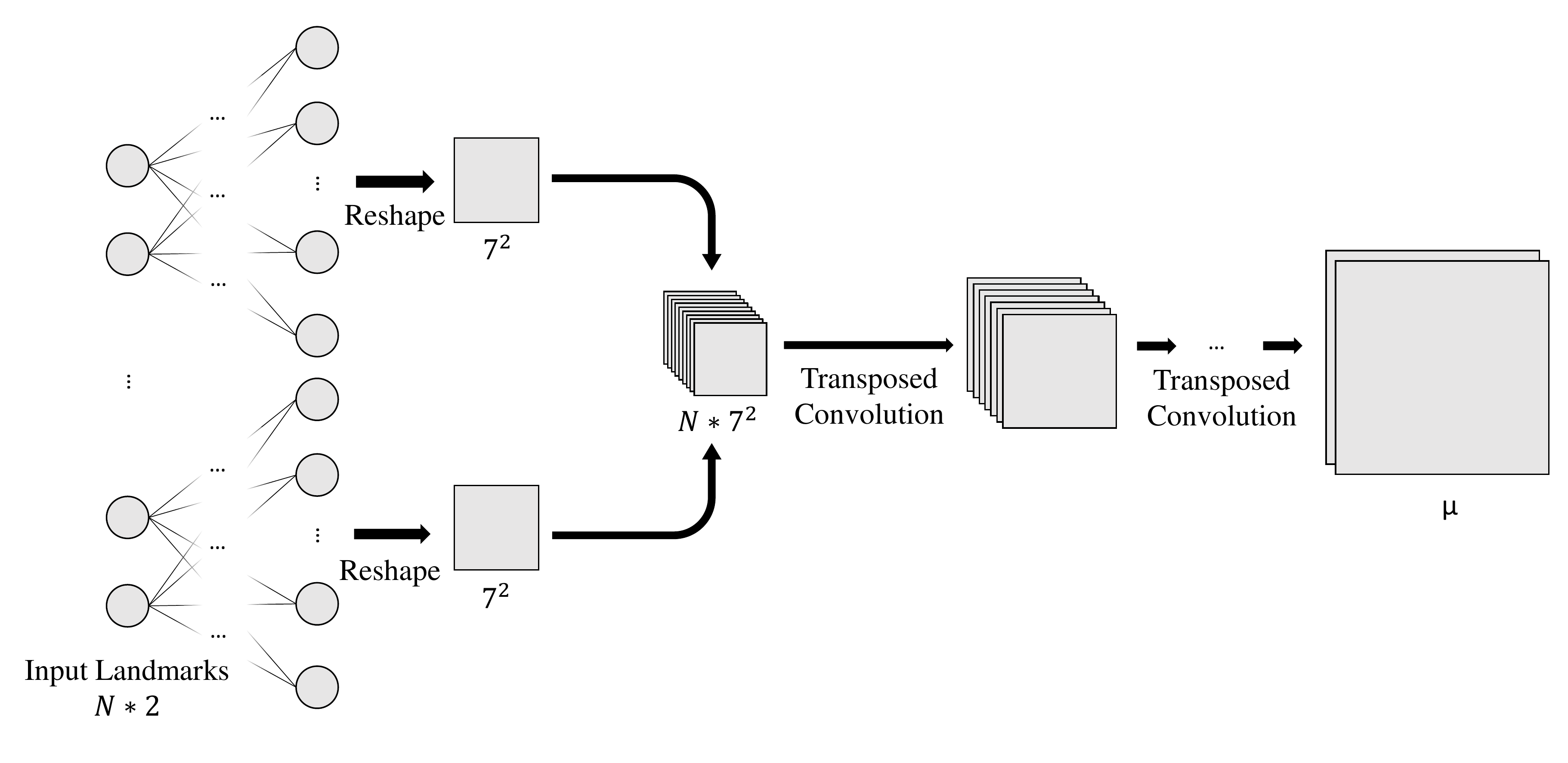}
    \caption{The structure of the CP-Net. We first process the input landmark locations through MLP and reshape them into separate channels before concatenation. Then we apply the transposed convolution and a tanh-type activation function to generate the Beltrami coefficients.}
    \label{MuGen}
\end{figure}

Initially, we use the MLP to process the starting points. We separate the points into groups and use separate MLPs to generate the feature map in the latent space. After that, we reshape the points into channels and perform a transposed convolution on them. Recall that for the DBS-Net, we adopt the bilinear interpolation for upsampling and do convolution right after that, as the information in the feature map is enough for us to get the output mapping. By contrast, the coefficient prediction task involves generating global information from certain sampled local information, while the reshaped landmark coordinates do not contain all information we need for the entire disk. To get a trainable upsampling for generating the Beltrami coefficients, we use the transposed convolution defined by padding and stride to get the global feature from the local feature. This gives a complex number output $\nu(T)$ for each triangle $T$. Finally, we would like to ensure that the quasi-conformal mappings subsequently generated by the output Beltrami coefficients of this network are bijective, i.e. there will be no folding in the resulting mappings. To achieve this, we follow the idea in~\cite{chen2021learning} and add the following tanh-type activation function $\mathcal{T}$ in the last layer:
\begin{equation}
    \mu(T) = \mathcal{T}(\nu(T)) = \tanh(|\nu(T)|) e^{i \arg(\nu(T))}.
\end{equation}
This ensures that $\|\mu\|_{\infty} < 1$ and hence the mappings are always bijective. The structure of the CP-Net is shown in Fig.~\ref{MuGen}.

\textbf{Loss Function}: Recall that $|\mu|$ measures the quasi-conformal distortion of a mapping and hence it is desired to be as small as possible. Therefore, we consider the following term:
\begin{equation}
\label{muloss}
    \mathcal{L}_\mu = \frac{1}{N}\sum^{N}_{i=1}\|\mu_i\|^2_2,
\end{equation}
where $N$ is the number of faces. Next, since $|\nabla\mu|$ controls the smoothness of the mapping, we consider the following term:
\begin{equation}
\label{musmooth}
    \mathcal{L}_{\nabla\mu} = \frac{1}{N}\sum^{N}_{i=1}\|\nabla\mu_i\|^2_2.
\end{equation}
Finally, the landmark mismatch error can be assessed by the following term:
\begin{equation}
\label{lmloss}
    \mathcal{L}_{\text{Landmark}} = \frac{1}{M}\sum^{M}_{i=1}\|p_i - \hat{p}_i\|^2_2,
\end{equation}
where $M$ is the number of landmarks, $p_i$ is the target location, and $\hat{p}_i$ is the network output, $i = 1, \dots, M$. By combining~\eqref{muloss}, \eqref{musmooth} and \eqref{lmloss}, we have the following loss function:
\begin{equation}
\label{RegLoss}
    \mathcal{L}_{\text{CP}} = \alpha\mathcal{L}_\mu + \beta\mathcal{L}_{\nabla\mu} + \gamma\mathcal{L}_{\text{Landmark}},
\end{equation}
where $\alpha,\beta,\gamma$ are nonnegative parameters. Note that for the training of the CP-Net, we need to have a pre-trained DBS-Net for the data generation and for producing the ultimate quasi-conformal mappings from the outputs of the CP-Net. This will be described in Section~\ref{sect:experiment}.

\section{Experiments}\label{sect:experiment}
In this section, we describe our implementation of the proposed framework and present the experimental results. 

\begin{figure}[t]
    \centering
    \includegraphics[width = \linewidth]{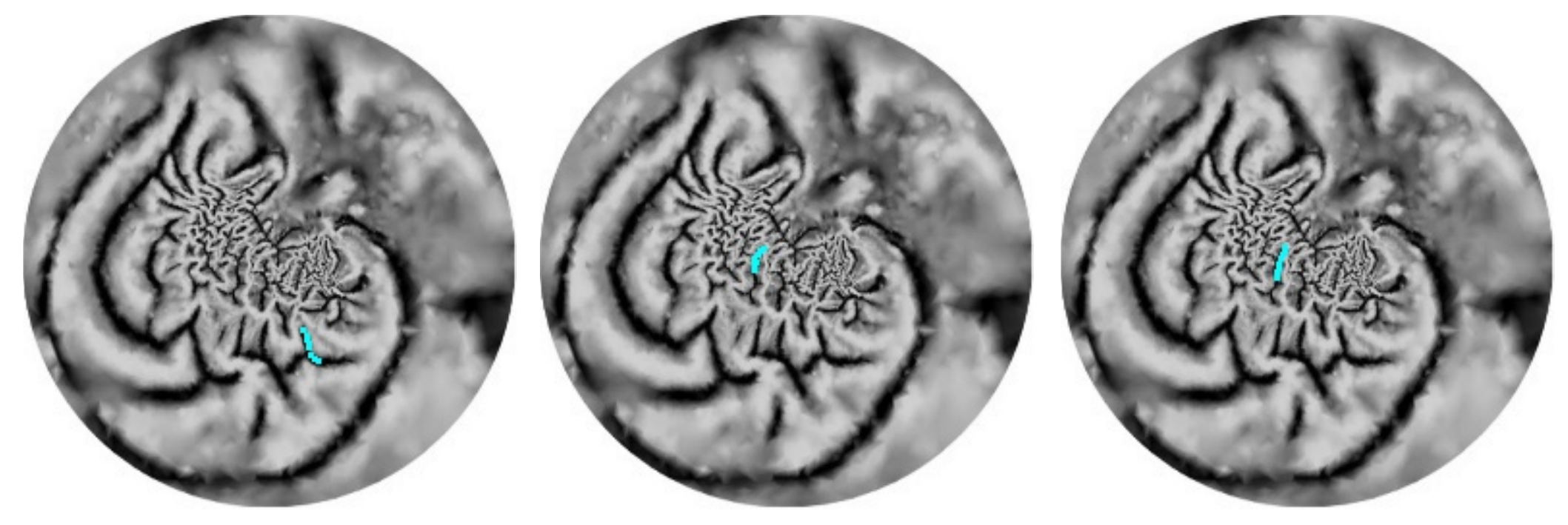}
    \caption{Some sample outputs from the landmark detection network (LD-Net). The blue curves show the detected landmarks, with the two endpoints are supplied as the input of the network.}
    \label{LandmarkDetectResults}
\end{figure}

\subsection{Training the Landmark Detection Network}
Our goal is to ensure that the LD-Net can produce landmark curves accurately along the sulci. To start with, we need to generate the training data for landmark detection. We first prepare multiple template brains and manually labelled landmarks along sulci as the prepared data. The brains are parameterized onto the unit disk using~\cite{choi2015fast}. Then, we randomly rotate and distort the brains by quasi-conformal maps and record the location of distorted landmarks. More specifically, images from the ImageNet~\cite{2009ImageNet} are randomly chosen to generate Beltrami coefficients in regular grid and produce the quasi-conformal maps. Note that here quasi-conformal maps are used so that the distorted brains will not have any folding, and hence this process will not lead to any disconnection of the sulci. We then take the starting and ending points of the sulci as well as the distorted brain as the network input to generate the detected landmark points and compare the result with the labels. Some examples are presented in Fig.~\ref{LandmarkDetectResults}, from which we can see that the sulci can be accurately extracted.

\subsection{Training the Disk Beltrami Solver Network}
\begin{figure}[t]
    \centering
    \includegraphics[width = 0.95 \linewidth]{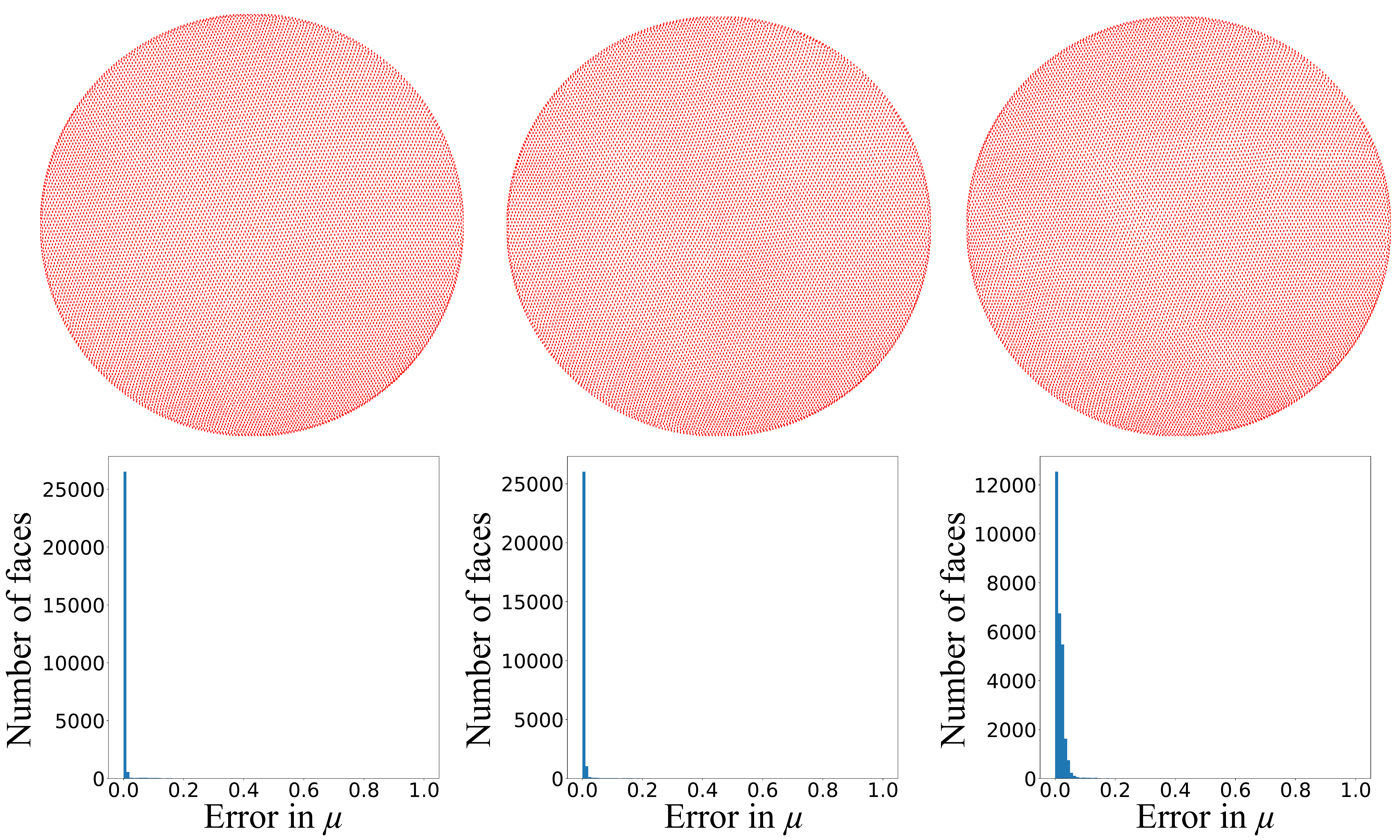}
    \caption{Three sample outputs of the DBS-Net with histograms of the error in the Beltrami coefficient $\mu$.}
    \label{BSnetResult}
\end{figure}

At the start of our training for the DBS-Net, we need sufficiently many Beltrami coefficient data on the disk for training. These data should be random and should contain as many scenarios as possible. To achieve this, we also use data from ImageNet~\cite{2009ImageNet} to augment the input Beltrami coefficients. In the augmenting process, we first convert all images into grayscale images and randomly select one as the real part and another as the imaginary part. We then smooth them and add random noise to them. The resulting two-channel image then serves as the input of DBS-Net, with the value for each pixel used for computing the LBS term in the loss function~\eqref{BSnetLoss}. Some example outputs of the DBS-Net are shown in Fig.~\ref{BSnetResult}.

\setlength{\tabcolsep}{0.5mm}{\begin{table*}[t]
\footnotesize
    \centering
\begin{tabular}{|c|c|c|c|c|c|}
    \hline
     Method & Mean $|\mu|$ & SD $|\mu|$ & Landmark Error & SD Landmark Error & Time (second) \\
     \hline 
     Our proposed method ($(\alpha, \beta, \gamma) = (1,1,10^4)$) & $1.87\times 10^{-2}$ & $2.78 \times 10^{-3}$  & $1.14 \times 10^{-2}$ & $2.59  \times 10^{-3}$ & 0.53\\
    \hline
    Our proposed method ($(\alpha, \beta, \gamma) = (1,1,20^5)$) & $4.67\times 10^{-2}$ & $6.01 \times 10^{-3}$ & $0.63 \times 10^{-2}$ & $2.20 \times 10^{-3}$ & 0.51\\
    \hline
    Optimized harmonic map ($\lambda = 0.1$) & $1.00\times 10^{-2}$ & $2.72  \times 10^{-3}$ & $4.10 \times 10^{-2}$& $12.1  \times 10^{-3}$ & 7.64 \\
    \hline
    Optimized harmonic map ($\lambda = 0.5$) & $2.90\times 10^{-2}$ & $8.37  \times 10^{-3}$ & $2.24 \times 10^{-2}$& $6.50 \times 10^{-3}$ & 14.96 \\
    \hline
    Optimized harmonic map ($\lambda = 1$) & $3.98\times 10^{-2}$ & $11.5 \times 10^{-3}$ & $1.19  \times 10^{-2}$ & $3.43  \times 10^{-3}$& 32.58 \\
    \hline
    Optimized harmonic map ($\lambda = 5$) & $5.16 \times 10^{-2}$ & $14.8 \times 10^{-3}$  & $0.51 \times 10^{-2}$ & $0.39  \times 10^{-3}$& 46.14 \\
    \hline
\end{tabular}
    \caption{A comparison between our framework and the landmark-constrained optimized harmonic mapping approach~\cite{choi2015flash,choi2015fast} for 24 brain cortical surfaces. Here, $\lambda$ is the soft landmark matching parameter in~\cite{choi2015flash}, and $\alpha, \beta, \gamma$ are parameters in~\eqref{RegLoss}.}
    \label{CompareWithDiskMap}
\end{table*}}

\begin{figure}[t!]
    \centering
    \includegraphics[width = 0.89\linewidth]{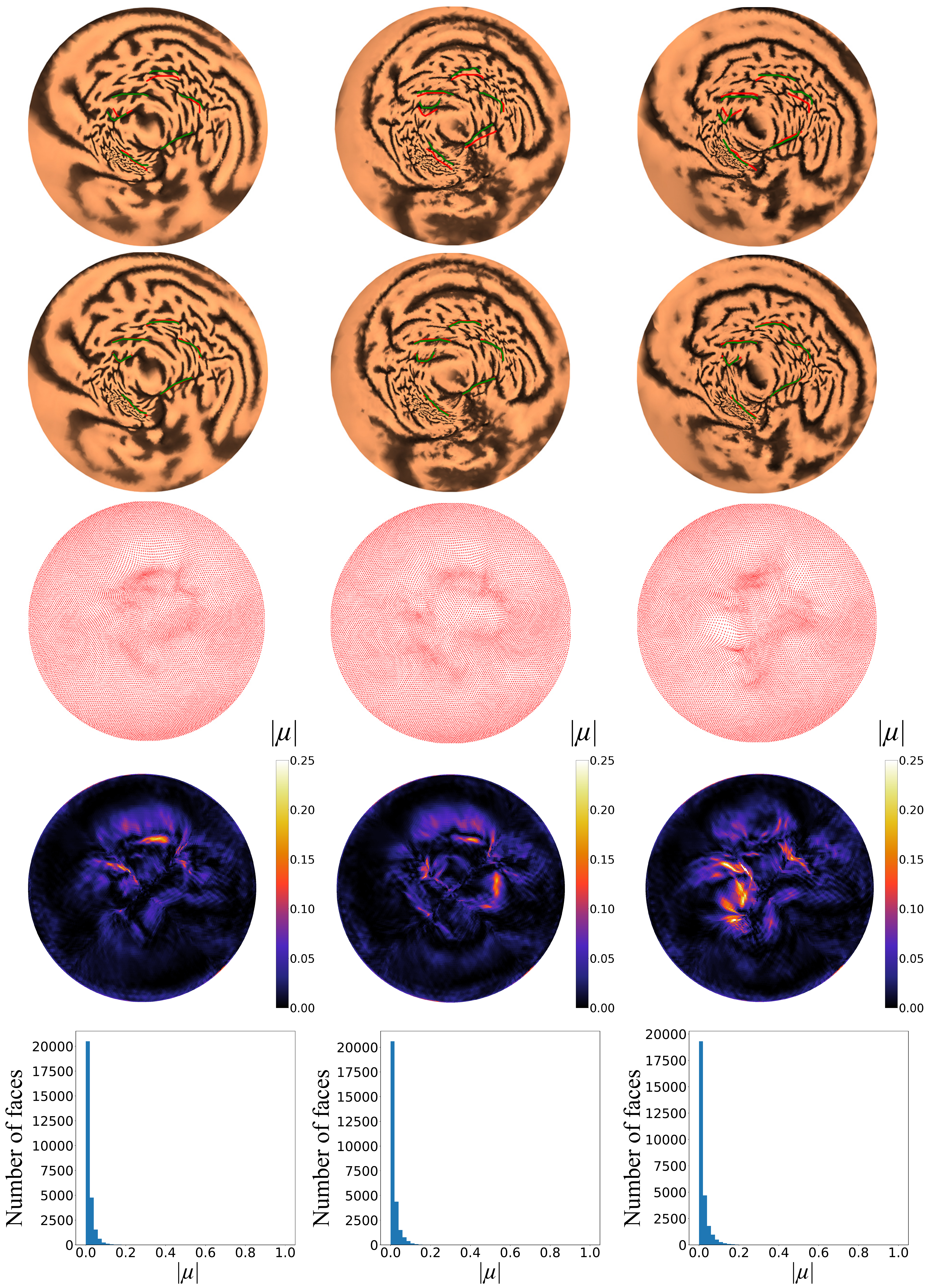}
    \caption{Some sample outputs of the proposed framework. Each column shows one experiment. The first row shows the initial disks produced using disk conformal parameterization~\cite{choi2015fast} color-coded by the surface curvature, where the green curves are the landmarks detected by the LD-Net and the red ones are the target locations of the landmarks. The second row shows the quasi-conformal maps on the disk produced by the CP-Net and the DBS-Net, from which it can be observed the landmarks are well-aligned. The third row shows an alternative representation of the mapping results. The fourth row shows the norm of the Beltrami coefficients $|\mu|$ on the unit disk. The fifth row shows the histograms of $|\mu|$.}
    \label{More-demo}
\end{figure}

\subsection{Training the Coefficient Prediction Network}
After getting a well-trained LD-Net for getting the landmark curves and a well-trained DBS-Net for generating quasi-conformal mappings based on any input Beltrami coefficients, the remaining step is to train the CP-Net to generate the Beltrami coefficients from the landmark constraints. 

To achieve this, we first need to prescribe the target positions of the landmarks. To generate the training data, we randomly choose two single-channel images and concatenate them to be two-channel data that acts as the Beltrami coefficient. We then use it as the input of the DBS-Net to generate a quasi-conformal mapping for perturbing the landmarks, and take the positions of the perturbed landmarks as the input of the CP-Net. This ensures that there is at least one admissible mapping without folding, i.e. $\|\mu\|_{\infty} < 1$. We use the distance between the resulting positions of the input points and the target positions to calculate the loss function~\eqref{RegLoss}. Note that~\eqref{RegLoss} also contains two terms for the norm of the Beltrami coefficients $\mu$ and $\nabla\mu$, which suppress the angle distortion.

\begin{figure}[t!]
    \centering
    \includegraphics[width = 0.9\linewidth]{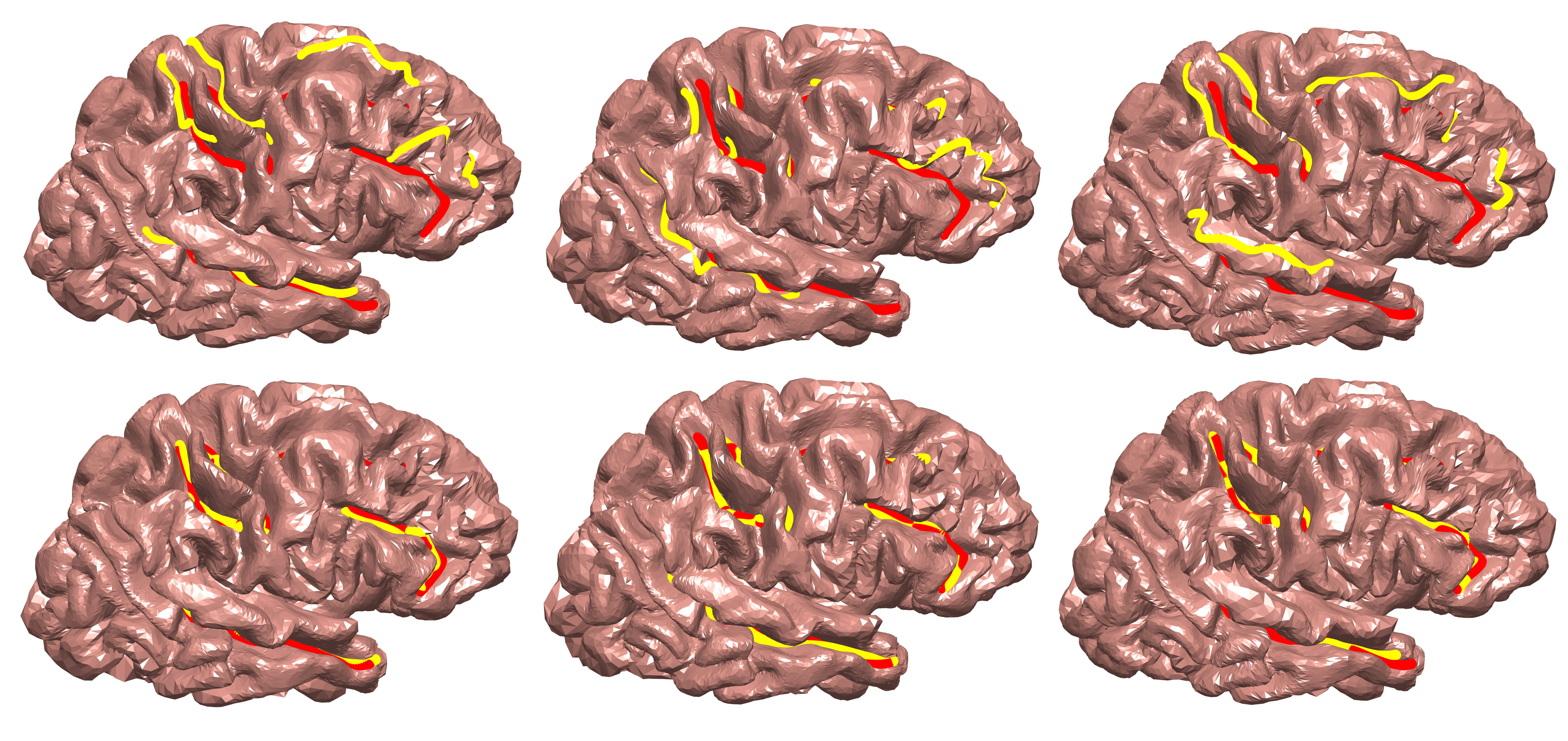}
    \caption{Examples of 3D brain registration results. The top row shows three registration results without using our proposed network, where the registered sulcal landmark curves are highlighted in yellow and the target landmarks are in red. The bottom row shows the registration results obtained by our framework, with the landmarks well-aligned.}
    \label{3D-demo}
\end{figure}

\subsection{Landmark-based Brain Cortical Surface Registration}
With all three proposed networks trained, we apply them for landmark-based brain cortical surface registration. 24 human brain cortical surfaces used in our experiments are reconstructed from MRI images from the Open Access Series of Imaging Studies (OASIS)~\cite{marcus2010open} using FreeSurfer~\cite{fischl2012freesurfer}.

We first conformally parameterize every brain onto the unit disk using~\cite{choi2015fast} and then apply the three networks for the landmark-based registration, where one brain from the dataset is selected as the control brain. As illustrated in Fig.~\ref{BrainDemo}, for every brain, we select the starting and ending points of each sulcus as the input and obtain the quasi-conformal map on the unit disk. We then apply the inverse map of the disk parameterization to map the disk back to the control brain in 3D. This completes the landmark-based brain registration. As shown in Fig.~\ref{More-demo} and Fig.~\ref{3D-demo}, the mappings match the sulcal landmarks accurately and possess very low quasi-conformal distortion. 

For a more quantitative analysis of all mapping results, we record the quasi-conformal distortion in terms of the norm of the Beltrami coefficients $|\mu|$, the landmark error and the computational time in Table~\ref{CompareWithDiskMap}. For comparison, note that landmark-constrained optimized harmonic mapping methods have been developed for registering genus-0 brain cortical surfaces~\cite{lui2007landmark,choi2015flash}, while the brain surfaces we handle in our framework are simply-connected open surfaces. Therefore, here we consider combining the disk conformal map method~\cite{choi2015fast} with the landmark-constrained optimized harmonic mapping method~\cite{choi2015flash} and compare the performance of this combined approach with our proposed framework. It can be observed that our proposed method achieves a significant improvement in the computational time by over 90\% when compared with the optimized harmonic map approach with different soft landmark matching parameters $\lambda$, while the quasi-conformal distortion and landmark error of the resulting registrations are comparable with those obtained using the optimized harmonic map approach. It is also noteworthy that by setting different training parameters $\alpha, \beta, \gamma$, we can achieve a lower landmark error or lower distortions depending on the goal of the task.

\begin{figure}[t]
    \centering
    \includegraphics[width = 0.9\linewidth]{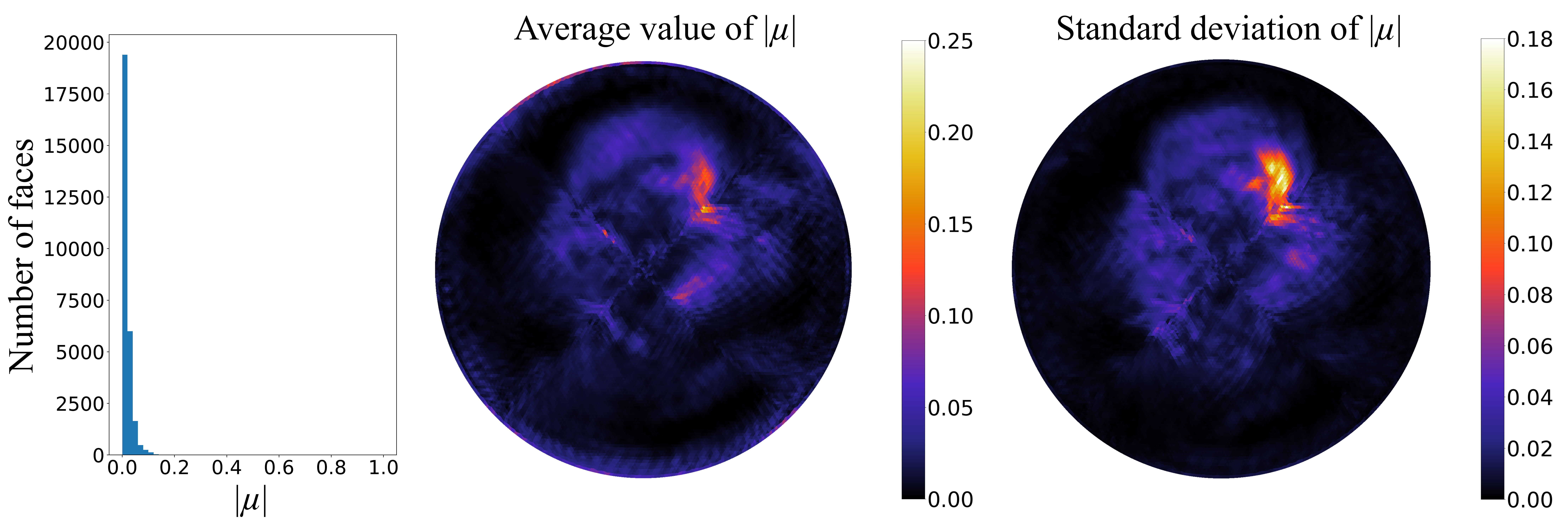}
    \caption{Analysis of the Beltrami coefficient $\mu$ of all mappings generated by our framework. The first plot shows the histogram of all face-based $|\mu|$ values. The second and third plots show the average value and the standard deviation of $|\mu|$ evaluated on each triangular face on the unit disk for all mappings.}
    \label{MuAnalysis}
\end{figure}

Also, in Fig.~\ref{MuAnalysis} we analyze the Beltrami coefficients $\mu$ of the resulting mappings for all brains. The histogram of $|\mu|$ shows that our framework achieves a minimal quasi-conformal distortion, with the majority of the $|\mu|$ values being close to 0. By taking the average of $|\mu|$ for all resulting mappings, we can again see that the quasi-conformal distortion is minimal in the entire unit disk. We further plot the standard deviation of $|\mu|$ evaluated on each triangular face, from which we can see a relatively large variation in $|\mu|$ near the landmarks, while the variation in $|\mu|$ far away from the landmarks is very small. 

\section{Conclusion}
In this paper, we have established a framework with three networks for the automatic landmark detection and surface registration of anatomical shapes with disk topology. Firstly, the LD-Net allows us to efficiently extract continuous feature landmarks from highly convoluted surfaces based on the endpoints of the desired feature curves and the surface curvature. Then, the CP-Net generates a Beltrami coefficient based on the detected landmarks. Finally, the DBS-Net, trained in unsupervised mode, produces a quasi-conformal mapping on the unit disk based on the generated Beltrami coefficients, which yields the surface registration result. The mappings produced by our framework possess low geometric distortion and preserve the bijectivity. Experimental results have demonstrated the effectiveness of our framework for brain surface registration. For future work, we plan to extend the framework for anatomical surfaces with other topologies.

\bibliographystyle{IEEEtran}
\bibliography{refer.bib}

\end{document}